\documentclass{article}
\usepackage{spconf,amsmath,graphicx}

\usepackage{enumitem}
\setlist{nosep, leftmargin=14pt}

\usepackage{mwe} % to get dummy images
\usepackage{hyperref}
\usepackage{url}
\usepackage{multirow}

\usepackage{xcolor}

\usepackage{amssymb}

\usepackage{graphicx}
\setkeys{Gin}{width=\linewidth, height=\textheight, keepaspectratio}

\usepackage{cite}
\usepackage{booktabs}
\title{{Cardiac Output Prediction from Echocardiograms: Self-Supervised Learning with Limited Data}}
\name{\begin{tabular}{c}
Adson Duarte$^{\star}$$^{\ddagger}$, Davide Vitturini$^{\star}$$^{\ddagger}$, Emanuele Milillo$^{\star}$, Andrea Bragagnolo$^{\dagger}$,\\ Carlo Alberto Barbano$^{\star}$, Riccardo Renzulli$^{\star}$, Michele Cannito$^{\star}$, Federico Giacobbe$^{\star}$,\\ Francesco Bruno$^{\star}$, Ovidio de Filippo$^{\star}$, Fabrizio D'Ascenzo$^{\star}$, Marco Grangetto$^{\star}$\thanks{$^{\ddagger}$These authors contributed equally to this work.}
\end{tabular}}
\address{$^{\star}$University of Turin, Italy
     $^{\dagger}$LINKS Foundation, Italy}
\begin{document}
%\ninept
%
\maketitle
\begin{abstract}
Cardiac Output (CO) is a key parameter in the diagnosis and management of cardiovascular diseases. However, its accurate measurement requires right-heart catheterization, an invasive and time-consuming procedure, motivating the development of reliable non-invasive alternatives using echocardiography. In this work, we propose a self-supervised learning (SSL) pretraining strategy based on SimCLR to improve CO prediction from apical four-chamber echocardiographic videos. The pretraining is performed using the same limited dataset available for the downstream task, demonstrating the potential of SSL even under data scarcity. Our results show that SSL mitigates overfitting and improves representation learning, achieving an average Pearson correlation of 0.41 on the test set and outperforming PanEcho, a model trained on over one million echocardiographic exams. Source code is available at \url{https://github.com/EIDOSLAB/cardiac-output}.
%Cardiac Output (CO) is a key parameter in the diagnosis and management of cardiovascular diseases. However, its accurate measurement requires right-heart catheterization, an invasive and time-consuming procedure, which motivates the development of reliable and non-invasive alternatives using echocardiography. In this work, we propose a self-supervised learning (SSL) pretraining strategy based on SimCLR to improve the prediction of CO from apical four-chamber echocardiographic videos. The pretraining is performed using the same limited dataset available for the downstream task, demonstrating the potential of SSL even when data is scarce for both pretraining and fine-tuning. Our results show that SSL helps mitigate overfitting and improves representation learning, achieving an average Pearson correlation of 0.41 for predicting CO under the test set, while obtaining better results than PanEcho, a model trained on over one million echocardiographic exams.
\end{abstract}
\begin{keywords}
self-supervised learning, contrastive learning, echocardiography, cardiac output
\end{keywords}
\section{Introduction}
\label{sec:intro}
Cardiac output (CO)~\cite{vincent2008understanding} is a hemodynamic parameter that indicates the ability of the heart to pump blood effectively throughout the body. %and being an important parameter in the diagnosis and management of cardiovascular diseases.
In clinical practice, CO is typically measured through right-heart catheterization~\cite{chen2020right}. Although this technique provides highly accurate measurements, it is also invasive for patients and time-consuming for clinicians. %, which limits its routine use.
In contrast, echocardiography~\cite{alsharqi2018artificial} is a non-invasive imaging technique that provides real-time visualization of cardiac motion, for example, through the apical four-chamber (A4C) view.
Considering that A4C acquisitions are much easier and faster to obtain than right heart catheterization, estimating CO directly from A4C could provide a safer and more accessible alternative to invasive hemodynamic assessment.

\begin{figure}
    \centering
    \includegraphics[width=\linewidth]{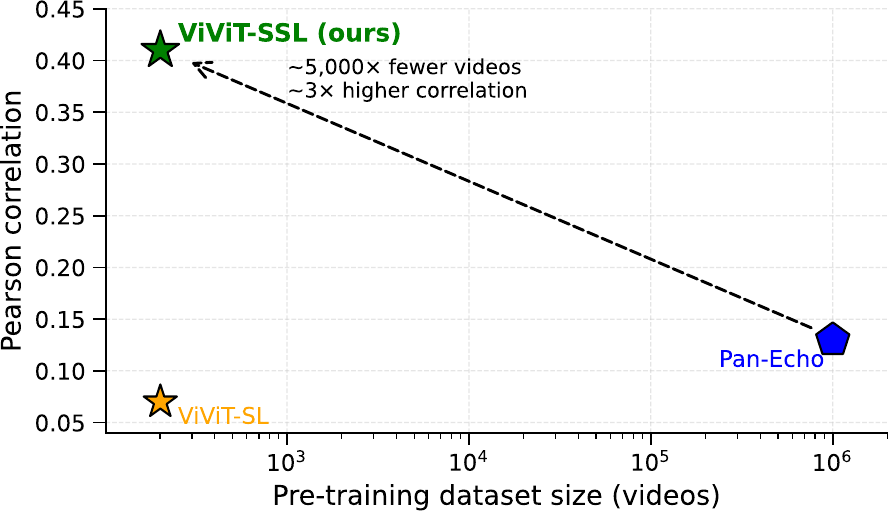}
    \caption{Test-set Pearson correlation versus pretraining dataset size (log scale, number of videos) for \textit{ViViT-SSL} (ours), \textit{PanEcho-R}, and \textit{ViViT-SL}.} % \textit{ViViT-SSL} achieves about $3{\times}$ higher correlation using $\sim\!5{,}000{\times}$ fewer pretraining videos than \textit{PanEcho-R}.
    \label{fig:teaser-data-efficiency}
\end{figure}

\iffalse
\begin{equation}
\label{eq:co}
    CO = HR {\times} SV
\end{equation}
\fi

Deep learning models require large amounts of annotated data to achieve strong generalization performance. However, in the medical domain, collecting such datasets is especially challenging since \textbf{\textit{(i)}} annotation is highly specialized and time-consuming, and \textbf{\textit{(ii)}} invasive ground-truth measurements, such as CO, are available only for a limited subset of patients.
In difficult tasks such as CO prediction, the scarcity of large datasets often leads supervised models to overfit training samples.
These limitations motivate the use of self-supervised learning (SSL) methods~\cite{krishnan2022self} that can utilize unlabeled data to learn robust representations.
In particular, contrastive approaches like SimCLR~\cite{chen2020simple} have shown promising gains in medical imaging~\cite{wang2023review}.
However, these gains typically rely on pretraining with large external datasets. A key question that remains unexplored is whether contrastive SSL can still offer benefits when pretraining is restricted to the same small dataset used for the downstream task.

In this work, we investigate the use of SimCLR as a pretraining strategy for a deep learning model aimed at predicting CO from A4C echocardiographic videos.
However, unlike related works that rely on large collections of unlabeled data for pretraining, we apply SSL using the same limited dataset available for the downstream task.
By doing so, we assess the benefits of SSL even when only a very small amount of data is available for both pretraining and fine-tuning. This is illustrated in Fig.~\ref{fig:teaser-data-efficiency}, where our method lies in the low-data corner yet still surpasses a large-scale supervised model, highlighting the value of SSL under data-scarce conditions.
%Moreover, we also investigate the impact of the batch size when employing contrastive SSL.
% TRAINING SIZE PART

\section{Related Works}
\label{sec:related_works}
% Several efforts have been made to advance automated cardiac assessment from echocardiograms. 
Deep learning has been extensively applied to echocardiography for tasks such as view classification, cardiac structure segmentation, and ejection fraction assessment~\cite{pandey2021deep, ouyang2020video, tokodi2023deep}.
With respect to CO estimation, prior work has predominantly leveraged non-echocardiographic modalities. For example,~\cite{xu2023improved} proposes a UNet-based model that predicts CO from photoplethysmography and arterial pressure waveforms, whereas~\cite{wang2025deep} employs a multimodal deep learning framework trained on electrocardiogram, seismocardiography, and body mass index data. In the echocardiography domain,~\cite{ufkes2023automatic} proposes a model that estimates CO using combined 5-chamber echocardiographic videos and Doppler measurements. Recently, Holste~\textit{et al.}~\cite{holste2025panecho}, presented PanEcho, a model trained on over one million examinations from roughly 20{,}000 patients in a fully supervised manner across a wide range of views, that predicts more than 30 clinically relevant cardiac measurements spanning the great vessels, chambers, and valves. Works such as these have substantially advanced automated cardiac function assessment from echocardiograms, however most remain centered on ejection fraction, and CO estimation typically relies on modalities or views other than the A4C view. Furthermore, existing approaches often depend on large external pretraining datasets and rarely investigate contrastive SSL. Accordingly, the contributions of this work are threefold: \textit{\textbf{(i)}} we propose a method for predicting CO directly from A4C echocardiography; \textit{\textbf{(ii)}} we evaluate contrastive SSL pretraining using only the small dataset available for the downstream task; and \textit{\textbf{(iii)}} we analyze the impact of batch size on SSL performance.

\section{Methodology}
\subsection{Pretraining}
\label{sec:pretraining}
We employ a self-supervised pretraining strategy based on contrastive learning to enhance CO prediction from echocardiographic videos in scenarios where only a limited amount of data is available, including the pretraining stage.
The approach can be applied to any video encoder.
During preliminary experiments, we tested several contrastive learning techniques, including exponentially weighted and kernelized supervised-contrastive objectives~\cite{barbano2023contrastive}, and ultimately selected SimCLR~\cite{chen2020simple}, which yielded the best results. SimCLR is a framework that constructs two stochastic augmentations of each input clip and learns representations by pulling together views of the same clip while pushing apart the others within the batch.
Let us assume that each input video $\mathbf{x}$ is mapped by the encoder $f(\cdot)$ into a latent representation $\mathbf{h} = f(\mathbf{x})$. On top of this representation, a small projection head $g(\cdot)$, implemented as a two-layer MLP, maps $\mathbf{h}$ into a space where the contrastive objective is applied, producing $\mathbf{z} = g(\mathbf{h})$. Following~\cite{chen2020simple}, the contrastive loss (NT-Xent) is computed between pairs of $\mathbf{z}$ vectors corresponding to different augmentations of the same video (positive pairs), encouraging similar representations for them, while pushing apart representations from other samples in the batch (negative pairs).
% After pretraining, the projection head $g(\cdot)$ is discarded, and the encoder representation $\mathbf{h}$ is retained for the downstream regression task.

The per-sample NT-Xent loss for anchor $i$~\cite{chen2020simple} is defined as

\begin{align}
\mathcal{L}_{i,\pi(i)}
&= - \log 
\frac{
    \exp\!\big(\mathrm{sim}(\mathbf{z}_i,\mathbf{z}_{\pi(i)})/\tau\big)
}{
    \sum_{\substack{k=1\\k\neq i}}^{2N}
    \exp\!\big(\mathrm{sim}(\mathbf{z}_i,\mathbf{z}_k)/\tau\big)
}.
\label{eq:contrastive_loss}
\end{align}

\noindent For each anchor \(i\), \(\pi(i)\) denotes the index of its corresponding augmented view, \(\{\mathbf{z}_m\}_{m=1}^{2N}\) denote the \(\ell_2\)-normalized embeddings of \(N\) clips, and \(\mathrm{sim}(\mathbf{z}_a,\mathbf{z}_b)=\mathbf{z}_a^\top \mathbf{z}_b\). Each clip has two augmented views, with temperature \(\tau>0\).

\subsection{Fine-tuning on the downstream task}
After SSL pretraining on the training split only, we freeze the video encoder and train a regression head on top of the learned representations $\mathbf{h}$. The SimCLR projection head $g(\cdot)$ is used only during pretraining and is discarded before regression. All architectural and optimization details are reported in Section~\ref{sec:results}.
This procedure can be applied to any video backbone, making the pretraining strategy flexible across architectures and suitable for data-scarce medical imaging scenarios.

\section{Experimental Results}
\label{sec:results}
\subsection{Dataset and preprocessing}
\label{subsec:dataset}
The dataset includes $268$ adult patients who underwent right-heart catheterization at Molinette and Normale di Pisa Hospitals. For each patient, A4C echocardiographic videos were acquired with Philips devices (iE33, CX50, EPIQ 7C), and CO was measured invasively during the procedure. Video resolutions range from $640{\times}360$ to $1024{\times}768$ pixels, with $16$–$181$ frames, frame rates of $10$–$75$ FPS, durations of $0.5$–$5$ seconds, and at least one complete cardiac cycle.

%\subsection{Preprocessing}
%\label{subsec:preprocessing}
%As a first step, patients with cardiac output values above 10 L/min were excluded, as such cases are easily identifiable according to clinical experts and predicting them offers limited clinical value. This resulted in a dataset of 336 videos, which were split into 201 for training, 68 for validation, and 67 for testing.
We randomly split the data into 201 training samples (75\%) and 67 test samples (25\%).
Each frame was center-cropped to focus on the cardiac region, resized to $224{\times}224$ pixels with bilinear interpolation, and normalized using ImageNet~\cite{deng2009imagenet} mean and standard deviation.
%Although originally stored in RGB, the ultrasound content is inherently grayscale, so all frames were converted to single-channel images.
All frames were then converted to single-channel images.

All videos were temporally resampled to 24 FPS. To ensure that each sequence captured at least one cardiac cycle, the first 32 frames (approximately 1.33 seconds) of each video were selected. For videos containing fewer than 32 frames, the available frames were centered, and the first and last frames were duplicated as needed at the beginning and end of the sequence to reach 32 frames.

\subsection{Model training}
All experiments use ViViT~\cite{arnab2021vivit} as the encoder. The model receives grayscale videos of shape $(N {\times} H {\times} W {\times} C)$ with $N{=}32$ frames, $H{=}W{=}224$, and $C=1$. We adopt the standard ViViT tubelet size and embedding dimension, producing a 768-dimensional feature vector. A regression head with one hidden layer (256 units, dropout 0.3) maps these features to a single CO prediction. This architecture is kept identical across all settings.
Performance is evaluated in terms of mean absolute error (MAE), Pearson correlation between predicted and ground-truth CO, and coefficient of determination ($R^2$) on the train and test sets.

%\subsubsection{Supervised learning}
\noindent\textbf{Supervised learning}: we begin by evaluating whether the representations learned by a  ViViT %in the pretraining performed by Google 
pretrained on generic video datasets~\cite{arnab2021vivit} could be effectively transferred to a medical application. As this approach proved insufficient, we proceeded to train ViViT end-to-end in a supervised manner, using data augmentation and regularization to reduce overfitting with our limited dataset.
%Subsequently, we investigated the feasibility of training ViViT end-to-end in a supervised manner, utilizing data augmentation and regularization to mitigate overfitting due to our limited dataset size.
In Table~\ref{tab:results}, \textit{ViViT-R (frozen)} and \textit{ViViT-SL (end-to-end)} refer, respectively, to supervised learning methods where only the regression head is trained while the ViViT encoder remains frozen, and where both the ViViT encoder and regression head are trained end-to-end.
Both methods show signs of overfitting: the MAE is consistently lower on the training set than on the test set, with an even more pronounced gap for the end-to-end approach. This behavior is expected, as training a model with $ \approx88\text{M} \text{ parameters} $ on a small dataset leads it to memorize the training data. %\mg{report the number of params of Vivit=88,549,761 with the projection head}
In contrast, \textit{ViViT-R (frozen)} overfits less, since only the regression layer is trained. However, its limited performance is also expected, given that ViViT was pretrained on out of domain data. The negative $R^2$ values for both methods in the test set confirm their failure to generalize, as they perform worse than simply predicting the mean.

\noindent\textbf{Self-supervised learning}: based on the unsatisfactory results obtained with supervised learning, we moved on to a SSL approach. As detailed in Section~\ref{sec:pretraining}, for the pretraining stage, we employ SimCLR and follow its standard data augmentation pipeline, including random resized cropping with horizontal flipping, strong color jitter with random grayscale conversion, and Gaussian blur. During pretraining, we use only the training set (201 samples) without labels, training the ViViT encoder for 500 epochs with Adam (learning rate $1{\times}10^{-5}$, weight decay $1{\times}10^{-3}$) and a step learning-rate schedule (decay by a factor of 0.1 at 33\% and 66\% of the total epochs). Two batch sizes (32 and 64) are tested across 4 NVIDIA A100 80GB GPUs, and three random seeds (40, 41, 42) are used to assess the robustness of the learned representations.
The \textit{ViViT-SSL-32} and \textit{ViViT-SSL-64} experiments correspond to the two batch size settings.
The results in Table~\ref{tab:results} show that contrastive SSL pretraining, even on a small dataset (201 samples), improves the representations of the encoder for CO regression. In fact, the best test performance (highlighted in bold) was obtained with \textit{ViViT-SSL-64}. Regarding the batch size, increasing it from 32 to 64 not only improved overall performance but also reduced variability in the MAE metric, where the standard deviation decreased from 0.33 to 0.004.
The modest Pearson correlation on the training set reflects the difficulty of predicting CO from A4C views, and training for more epochs tended to overfit, which we empirically mitigated by limiting the number of training epochs for the downstream regression task.

%% Tables place here just temporarily
\begin{table*}[t]
    \caption{Comparison of methods for cardiac output estimation from four-chamber view echocardiography videos.}
    \label{tab:results}
    \centering
    \small
    \begin{tabular}{lcccccc}
        \toprule
        \multirow{2}{*}{\textbf{Method}} & \multicolumn{2}{c}{\textbf{MAE}} & \multicolumn{2}{c}{\textbf{Pearson}} & \multicolumn{2}{c}{$\mathbf{R}^2$}\\
        \cmidrule(lr){2-3} \cmidrule(lr){4-5} \cmidrule(lr){6-7}
        & \textbf{Train} & \textbf{Test} & \textbf{Train} & \textbf{Test} & \textbf{Train} & \textbf{Test} \\
        \midrule
        Baseline (Mean) & $1.17$ & $1.17$ & - & - & - & - \\
        \midrule
        ViViT-R (frozen) & $0.79_{\pm0.01}$ & $1.12_{\pm 0.01}$ & $0.24_{\pm0.02}$ & $0.24_{\pm0.02}$ & $0.02_{\pm0.01}$ & $-0.04_{\pm 0.02}$ \\
        ViViT-SL (end-to-end) & $0.07_{\pm 0.01}$ & $1.30_{\pm 0.02}$ & $0.51_{\pm 0.02}$ & $0.02_{\pm 0.02}$ & $0.25_{\pm 0.02}$ & $-0.31_{\pm 0.03}$ \\
        \midrule
        ViViT-SSL-32 & $1.61_{\pm 0.03}$ & $1.11_{\pm 0.33}$ & $0.39_{\pm 0.01}$ & $0.34_{\pm 0.01}$ & $0.13_{\pm 0.01}$ & $0.05_{\pm 0.03}$ \\
        ViViT-SSL-64 & $1.08_{\pm 0.004}$ & $\boldsymbol{1.05_{\pm 0.004}}$ & $\boldsymbol{0.48_{\pm 0.02}}$ & $\boldsymbol{0.41_{\pm 0.01}}$ & $\boldsymbol{0.22_{\pm 0.01}}$ & $\boldsymbol{0.12_{\pm 0.01}}$ \\
        \midrule
        PanEcho-R & $\boldsymbol{0.99_{\pm 0.04}}$ & $1.14_{\pm 0.02}$ & $0.46_{\pm 0.03}$ & $0.13_{\pm 0.01}$ & $0.18_{\pm 0.01}$ & $-0.11_{\pm 0.02}$ \\
        %giving the variability of the results (mean and standard deviation)
        \bottomrule
    \end{tabular}
\end{table*}
%Table~\ref{tab:results} shows the obtained results for each method, including the Mean Absolute Error (MAE), Pearson's correlation coefficient, and $\mathbf{R}^2$.
%The \textit{Baseline (Mean)} method indicates the obtained MAE when simply predicting the mean value, serving as a reference to assess the performance of each method.
\noindent\textbf{Comparison with PanEcho}: we compare our SSL approach with PanEcho~\cite{holste2025panecho}, a model trained in a fully supervised manner on $\approx1\text{M samples}$, including A4C echocardiography videos. While PanEcho was not trained to predict CO, we evaluate its learned representations, which are in the same domain as ours, by freezing its encoder and training the same regression head used in our experiments. This allows a direct comparison of how representations learned from large-scale supervised data transfer to our CO prediction task, relative to those learned via SSL on a small dataset (only 201 samples).
In Table~\ref{tab:results}, the \textit{PanEcho-R} results show that despite PanEcho being trained in a fully supervised manner on approximately $\approx 1\text{M samples}$ from the same domain as ours, its learned representations do not transfer effectively to the CO prediction task. Our SSL approach (\textit{ViViT-SSL-64}) consistently outperforms PanEcho on the test set across all evaluation metrics, indicating the effectiveness of SSL even when only a very small dataset is available for the pretraining stage.

\begin{figure}[t]
    \centering
    \includegraphics[width=0.8\linewidth]{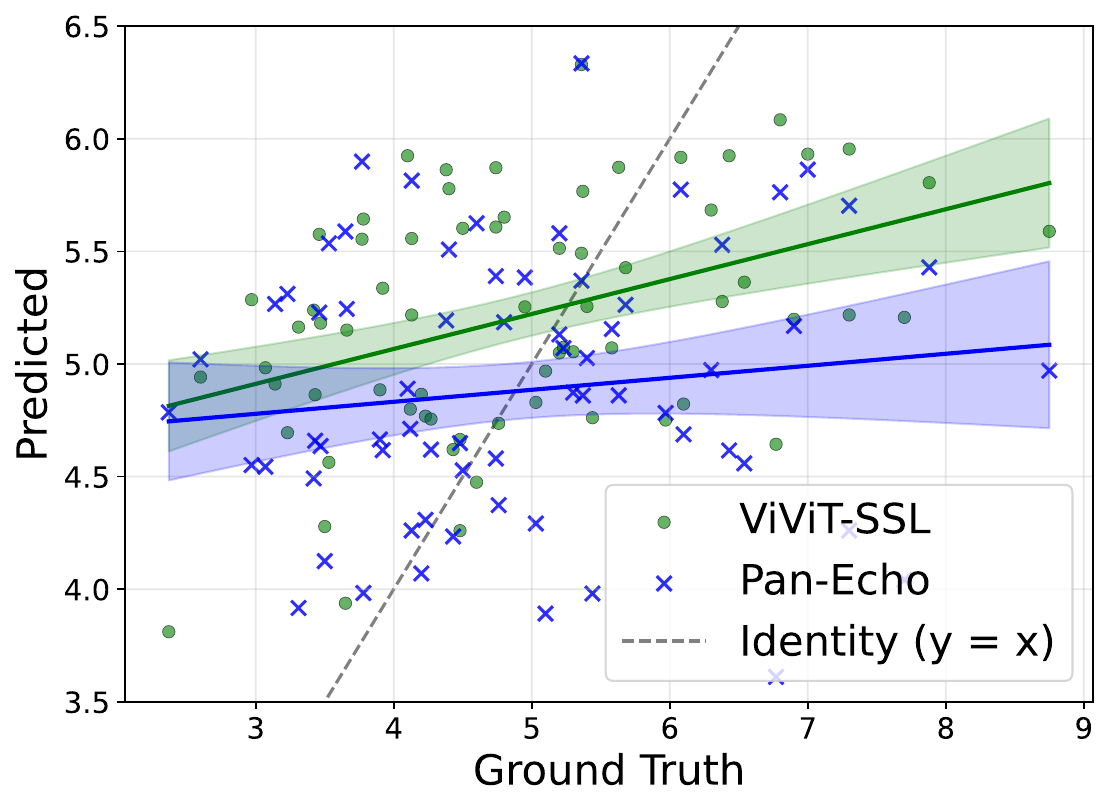}
    \caption{Predicted versus ground-truth CO on the test set for \textit{ViViT-SSL} (green) and \textit{PanEcho-R} (blue). Solid lines show linear regression fits; shaded regions denote 90\% confidence intervals.}
    \label{fig:regression_scatter}
\end{figure}

These differences are also visible in the regression plots in Fig.~\ref{fig:regression_scatter}. \textit{ViViT-SSL} (green) yields a positively sloped fit with a narrow confidence band and points distributed around the line across the CO range, indicating that the model captures the underlying variability. In contrast, \textit{PanEcho} (blue) exhibits an almost flat fit with a wide confidence interval and predictions fluctuating around the mean CO, consistent with a weak dependence on the true CO and with the low Pearson correlation and negative $R^2$ reported in Table~\ref{tab:results}.
% TRAINING SIZE PART
Fig.~\ref{fig:teaser-data-efficiency} summarizes the trade-off between accuracy and data scale, where \textit{ViViT-SSL} reaches about $3{\times}$ higher Pearson correlation than \textit{PanEcho-R} despite relying on roughly $5{,}000{\times}$ fewer pretraining videos.

\section{Conclusion}
\label{conclusion}
Our results demonstrate that SSL can be an effective strategy for challenging medical imaging tasks, such as predicting CO from A4C views, even when the available dataset is extremely limited.
We also observed that increasing the batch size during SSL stabilizes the learned representations and improves downstream performance.
In comparison, a large-scale supervised model trained on one million echocardiography videos (PanEcho) does not transfer as effectively to this task, highlighting the potential of SSL to extract task-relevant features from small datasets.
Despite these promising findings, our study is limited to a single dataset, a specific imaging view, and a single regression task, and further work is needed to evaluate the generalizability of this approach across different datasets, modalities, and clinical endpoints.
%Future research could explore alternative SSL pretext tasks, fine-tuning strategies, and applications to broader medical imaging problems, aiming to maximize the utility of limited labeled data in clinical practice.

\noindent\textbf{{Compilance with Ethical Standards}}
%\section{Compilance with Ethical Standards}
This study was performed in accordance with the Declaration of Helsinki and approved by the institutional ethics committee. All patient data were anonymized prior to processing.


\bibliographystyle{IEEEbib}
\bibliography{refs}

\end{document}